\documentclass{bmvc2k}
\usepackage{graphicx} 
\usepackage{tabu}
\usepackage{multirow}

\title{Few-Shot Anomaly Detection with Adversarial Loss for Robust Feature Representations}

\addauthor{Jae Young Lee}{mcneato@kaist.ac.kr;mcneato@etri.re.kr}{1,3}
\addauthor{Wonjun Lee}{wjdnjswns@ust.ac.kr}{2}
\addauthor{Jaehyun Choi}{chlwogus@kaist.ac.kr;chlwogus@etri.re.kr}{1,3}
\addauthor{Yongkwi Lee}{glory1210@etri.re.kr}{3}
\addauthor{Young Seog Yoon}{isay@etri.re.kr}{3}

\addinstitution{
Korea Advanced Institute of \\Science and Technology (KAIST), \\
Daejeon, South Korea
}
\addinstitution{
University of Science and \\Technology (UST),\\
Daejeon, South Korea
}
\addinstitution{
Electronics \& Telecommunications \\Research Institute (ETRI),\\
Daejeon, South Korea
}

\runninghead{J. Y. Lee et al.}{Few-shot anomaly detection with adversarial loss}

\begin{document}
\maketitle

\begin{abstract}
Anomaly detection is a critical and challenging task that aims to identify data points deviating from normal patterns and distributions within a dataset. 
Various methods have been proposed using a one-class-one-model approach, but these techniques often face practical problems such as memory inefficiency and the requirement of sufficient data for training. 
In particular, few-shot anomaly detection presents significant challenges in industrial applications, where limited samples are available before mass production. 
In this paper, we propose a few-shot anomaly detection method that integrates adversarial training loss to obtain more robust and generalized feature representations. 
We utilize the adversarial loss previously employed in domain adaptation to align feature distributions between source and target domains, to enhance feature robustness and generalization in few-shot anomaly detection tasks.
We hypothesize that adversarial loss is effective when applied to features that should have similar characteristics, such as those from the same layer in a Siamese network's parallel branches or input-output pairs of reconstruction-based methods.
Experimental results demonstrate that the proposed method generally achieves better performance when utilizing the adversarial loss.
\end{abstract}

\section{Introduction}
\label{sec:intro}
Anomaly detection is a challenging task that involves identifying data points that deviate from the normal patterns and distributions within a dataset. 
Due to the complexity and diversity of data, various anomaly detection methods have been proposed using a one-class-one-model approach, where only normal samples are used for training since abnormal data are infrequently observed. 
However, this approach poses practical problems as each class requires its own dedicated model, which results in a memory-inefficient model. 
Furthermore, few-shot anomaly detection (FSAD) is a practical and challenging problem, particularly in the industrial sector during the product development process when there are insufficient samples available before mass production.

In the existing literature, anomaly detection methods are generally developed under the assumption that sufficient data is available for training. 
These methods employ a one-class learning paradigm, evolving into a one-class-one-model paradigm with various techniques being proposed.  
Major branches of these techniques include reconstruction-based methods \cite{superpixel, draem, divide, dsr, Anogan, ganomaly, anoseg} and embedding similarity-based methods \cite{mahalanobisAD, SPADE, padim, patchcore}, which have led to the development of distinct architectures. 
The reconstruction-based methods employ generative models to learn the distribution of normal samples, reconstructing normal images and comparing them with input samples during inference time to determine anomalies. 
On the other hand, the embedding similarity-based methods use pre-trained encoders to extract features and make a feature distribution for normal samples. 
During inference, the features of input samples are compared with the normal feature distribution.

Recently, these techniques have expanded the scope of anomaly detection methods.
UniAD \cite{uniad} has expanded the one-class-one-model paradigm to multi-class anomaly detection.
In the UniAD model, they extract features from the encoder and mask them for reconstruction. 
Similar to the embedding similarity-based methods, UniAD extracts features from encoders and reconstructs masked features like the reconstruction-based methods.
RegAD \cite{regad} adopts a Siamese network structure with an additional registration step and an encoder-predictor architecture. 
RegAD has expanded the domain of FSAD, offering a more versatile solution to the challenges faced in this field.
These methods use a single loss function suitable for each method to cluster features in high-dimensional spaces, enabling features from normal samples to form clusters. 
However, in FSAD, it is crucial to effectively utilize the limited samples, especially during the initial product development stages before mass production, when an abundance of samples is not accessible. 

In this paper, we propose integrating a type of adversarial loss into FSAD tasks to further optimize the use of limited samples and obtain more robust and generalized feature representations, which could potentially improve anomaly detection performance. 
The adversarial loss has been employed in domain adaptation to align feature distributions between the source and target domains. 
By incorporating an additional discriminator and confusing it from differentiating between the labels of the source and target, the primary model generates robust and generalized features, even for target domain data. 
In FSAD, we hypothesize that the adversarial loss can enhance feature robustness and generalization when applied to different features that should possess similar characteristics, such as features from the same layer in a Siamese network's parallel branches and input-output pairs of reconstruction-based methods. 
Experiments with MVTec \cite{mvtec} and DAGM \cite{dagm2007} datasets demonstrate that the proposed method generally achieves better performance when utilizing the adversarial loss.

\section{Related Work}
\label{sec:Related work}
\subsection{Anomaly Detection}
Previously, anomaly detection was mainly performed using traditional statistical methods \cite {IQR, statistical, statistical2}, but with the advancement of deep learning, various deep learning-based anomaly detection methods have been proposed.
In this section, we summarize the reconstruction-based methods, embedding similarity-based methods, and the existing few-shot methods, which represent the main types of anomaly detection techniques.

The reconstruction-based methods use generative models such as variational auto-encoder (VAE) \cite{VAE}, generative adversarial network (GAN) \cite{GAN}, and so on to learn the distribution of normal samples. 
The reconstruction-based methods \cite{superpixel, draem, divide, dsr, Anogan, ganomaly, anoseg} detect anomalies by assuming that the model trained on normal samples will reconstruct normal samples well while reconstructing poor abnormal samples.
Several methods have been proposed for anomaly detection using an auto-encoder. 
Superpixel Masking And Inpainting (SMAI) \cite{superpixel} is an unsupervised approach that employs superpixel segmentation and inpainting to identify and localize abnormal regions in images by comparing the original and reconstructed mask areas.
DRÆM \cite{draem} treats surface anomaly detection as a discriminative problem, learning a joint representation of anomalous images and their anomaly-free reconstructions. 
This method allows for direct anomaly localization without complex post-processing and can be trained with simple simulations.
Divide-and-Assemble \cite{divide} adjusts the reconstruction capability for both normal and abnormal samples by varying the granularity of division on feature maps. 
It incorporates a multi-scale block-wise memory module in an auto-encoder network and introduces adversarial learning to enhance subtle anomaly detection.
DSR \cite{dsr} proposes a dual subspace re-projection network for surface anomaly detection, utilizing an auto-encoder-based approach to improve the discriminative capability between normal and abnormal samples, ultimately boosting anomaly detection performance.
Also, there are several methods using GAN for anomaly detection.
AnoGAN \cite{Anogan} trains a GAN on normal data samples and measures the dissimilarity between input samples and generated samples to identify anomalies. 
GANomaly \cite{ganomaly} builds on this by utilizing a semi-supervised framework with adversarial training, focusing on learning latent representations of normal data for efficient anomaly detection. 
Anoseg \cite{anoseg} further advances GAN-based techniques by introducing a self-supervised anomaly segmentation network, highlighting the potential of GAN for anomaly segmentation tasks. 

The embedding similarity-based methods \cite{mahalanobisAD, SPADE, padim, patchcore} extract feature distribution from normal samples using the pre- trained model trained on a large dataset. 
In inference time, the input samples are classified into normal and abnormal by comparing the features of the normal samples and the input sample.
MahalanobisAD \cite{mahalanobisAD} extracts features from a pre-trained model and computes the Mahalanobis distance between these features and the mean of normal data features. 
SPADE \cite{SPADE} extracts features from multiple layers of a pre-trained network and builds a pyramid of image patches for each layer. Then, it searches for the most similar patches in the reference set using k-nearest neighbors (KNN).
PaDiM \cite{padim} obtains features from the last three layers of a pre-trained model and computes the mean and covariance for each patch.
PatchCore \cite{patchcore} constructs a set of prototypes from the features extracted using a pre-trained network, and during inference, it measures the distance between these prototypes and the features of input images to identify anomalies.

Recently, these methods are combined and the abilities of the methods are enlarged to handle practical problems such as few-shot \cite{TDG, differnet,regad} and multi-class anomaly detection \cite{uniad}.
Especially in FSAD, TDG \cite{TDG} utilizes a hierarchical transformation-discriminating generative model that learns multi-scale patch distributions and improves model representation through image transformations. 
DifferNet \cite{differnet} employs CNNs and normalizing flows for feature extraction and density estimation, enabling anomaly detection based on likelihood in images. 
RegAD \cite{regad} introduces a category-agnostic feature registration method, leveraging a feature registration network and Siamese network for effective anomaly detection across diverse image categories with limited data.
In this paper, RegAD \cite{regad} and UniAD \cite{uniad} are utilized as baselines to show the effectiveness of the proposed method in FSAD. 

\subsection{Adversarial Loss in Domain Adaptation}
Domain adaptation aims to train a model that works equivalently well on the target domain with limited access to data, labeled source domain, and unlabelled target domain.
In domain adaptation, DANN \cite{DANN_C, DANN_J}
aligns feature distributions of the source and target domain. 
DANN introduces an adversarial training objective, where a domain discriminator is trained to differentiate between the source and target domain features, while the main model is optimized to generate features that are indistinguishable from the discriminator.
The gradient of the adversarial training objective flows from the discriminator to the main model, making features of the main model more robust and domain-invariant, thus improving the generalization on the target domain. 
The loss function (adversarial training objective) proposed by DANN is utilized in various fields such as semantic segmentation \cite{advent}, object detection \cite{DA_Faster_RCNN}, facial expression recognition \cite{ExprADA}, pose estimation \cite{Pose_DA}, and so on. 
These methods adopt the loss function proposed by DANN for domain adaptation. 
In this paper, we integrate a type of adversarial training objective to train the discriminator.
In contrast to prior methods, we utilize the adversarial loss between branches of a Siamese network or input-output pairs of reconstruction-based methods with features that should have similar characteristics.

\section{Proposed Method}
\subsection{Problem Formulation}
For the problem of FSAD \cite{regad}, a training set is composed solely of normal samples across $n$ categories, denoted as $D_{\text{train}} = \bigcup_{i=1}^{n} D_i$, where each subset $D_i$ includes normal samples from categories $c_i$ (for $i = 1, 2, ..., n$). 
Our goal is to develop a category-agnostic anomaly detection model using this training data.
During the testing phase, we use an image from a target category $c_t$, where $c_t$ does not belong to the set of known categories, i.e., $t \notin {1, 2, ..., n}$. 
Along with this image, $K$ normal samples from the target category $c_t$ are provided. 
The task for the trained anomaly detection model is to evaluate whether this image is anomalous or normal.

\subsection{Loss Function and Training Process}
In the proposed method, we consider a main model $M$, which could be applicable for anomaly detection such as RegAD, UniAD, or any suitable model.
The existing loss function used in these models is denoted by $\mathcal{L}_M$. 
Our approach introduces an auxiliary network, the discriminator $D$, which is trained alongside $M$ using an additional adversarial loss.
Adversarial learning in our method comprises two parts: 1) training the discriminator and 2) fooling the discriminator. 
For adversarial loss, we choose $f_0$ and $f_1$ as a pair of features within $M$ that have similar characteristics and have the same dimensions. 
For training $D$, our objective is to correctly classify $f_0$ and $f_1$ as belonging to distinct classes, labeled as 0 and 1 respectively. 
As the first part, the corresponding total loss function $\mathcal{L}_{DT}$ for the discriminator is thus formulated as:
\begin{equation}
\mathcal{L}_{DT} = \mathcal{L}_{D}(f_0, 0) + \mathcal{L}_{D}(f_1, 1),
\end{equation}
\noindent where $\mathcal{L}_{D}$ represents the cross-entropy loss function with appropriate labels.
Simultaneously, through training $M$, we aim to enhance its discriminative capability between normal and anomalous samples at the inference time. 
For this purpose, we define the total loss function $\mathcal{L}_{MT}$ as a combination of the original model loss function $\mathcal{L}_M$ and the adversarial loss $\mathcal{L}_{D}$ applied to $f_0$. 
We use the cross-entropy loss function but with the intention to misguide the discriminator, which is originally assigned a label of 0 for this feature.
As the second part, the total loss function $\mathcal{L}_{MT}$ for the main model is expressed as:
\begin{equation}
\mathcal{L}_{MT} = \mathcal{L}_M + \mathcal{L}_{D}(f_0, 1).
\end{equation}
\noindent Through this formulation, our proposed method strives to enhance the performance of anomaly detection models by boosting the generalization power of the features they extract.

With these loss functions, the training process for the proposed method is divided into two main steps: updating the parameters of the main model $\theta_M$ and updating the parameters of the discriminator $\theta_D$.
In the process of updating $\theta_M$, the weights of the main model are updated using the gradient of the total loss function $\mathcal{L}_{MT}$ with respect to $\theta_M$. This can be represented as:
\begin{equation}
\theta_M = \theta_M - \eta \nabla_{\theta_M} \mathcal{L}_{MT},
\end{equation}
\noindent where $\eta$ is the learning rate.
After updating $\theta_M$, in the process of training $D$, the model weights of the discriminator $\theta_D$ are updated using the gradient of the discriminator loss function $\mathcal{L}_{DT}$ with respect to $\theta_D$. 
It can be expressed as:
\begin{equation}
\theta_D = \theta_D - \eta \nabla_{\theta_D} \mathcal{L}_{DT}.
\end{equation}

\subsection{Integration into Existing Methods}
\begin{figure*}[!t]
    \centering
    \includegraphics[width=12.5cm]{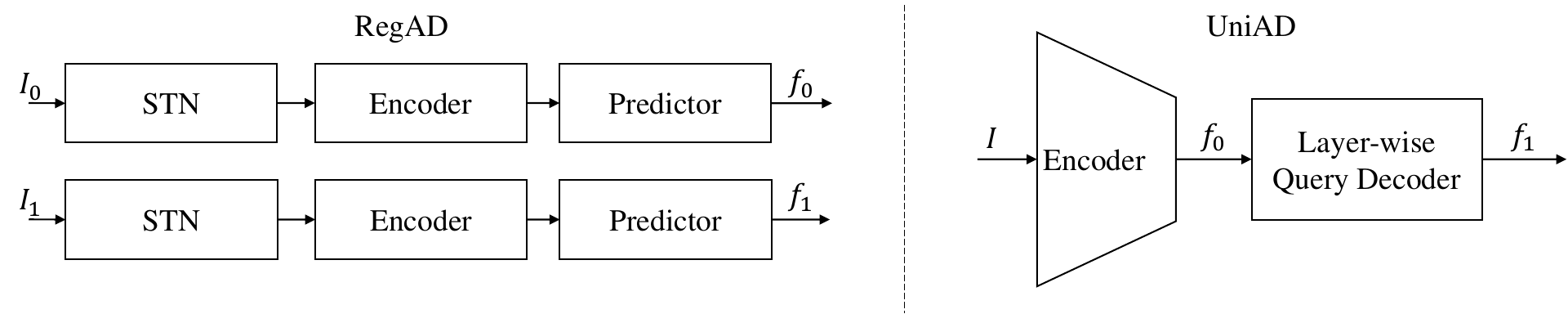}
    \caption{Architectures of RegAD and UniAD with features $f_0$ and $f_1$ used in the adversarial training. }
    \label{fig:architecture}
\end{figure*}

Figure \ref{fig:architecture} shows the architecture of both the RegAD and UniAD models with the features $f_0$ and $f_1$ used in the adversarial training.
In the RegAD model, a Siamese network architecture is employed, where each branch uses an image from a pair of images, denoted as $I_0$ and $I_1$. 
These images are randomly selected from the same category within the training set. 
Each branch is composed of three components: a spatial transformer network (STN), an encoder, and a predictor. 
After processing through these components, the output features from the predictor in each branch, represented as $f_0$ and $f_1$, should exhibit similar characteristics due to the nature of the Siamese architecture.
In the adversarial training, we use the outputs of each branch as $f_0$ and $f_1$.
The UniAD model consists of an encoder and a layer-wise query decoder. 
The encoder's output features are fed into the layer-wise query decoder in conjunction with neighborhood masking. 
The function of the layer-wise query decoder is to reconstruct the masked input features as non-masked features. 
Therefore, in the context of UniAD, the input-output pairs of the layer-wise query decoder should have similar characteristics. 
Hence, the input and output of the layer-wise query decoder are chosen as $f_0$ and $f_1$, respectively.

\section{Experiments}

\begin{table*}[t!]
\centering
\resizebox{\linewidth}{!}
{
\begin{tabular}{crrrrrrrrrrrrrrrr|r}
\hline\hline
&& \multicolumn{15}{c}{Category} \\ \cline{3-18}
\multicolumn{1}{c}{Shot} &\multicolumn{1}{c}{Method} & \multicolumn{1}{c}{\rotatebox[origin=r]{90}{Bottle}}		& \multicolumn{1}{c}{\rotatebox[origin=r]{90}{Cable}}		& \multicolumn{1}{c}{\rotatebox[origin=r]{90}{Capsule}}		
			  & \multicolumn{1}{c}{\rotatebox[origin=r]{90}{Carpet}}		& \multicolumn{1}{c}{\rotatebox[origin=r]{90}{Grid}}		& \multicolumn{1}{c}{\rotatebox[origin=r]{90}{Hazelnut}} 		
			  & \multicolumn{1}{c}{\rotatebox[origin=r]{90}{Leather}}		& \multicolumn{1}{c}{\rotatebox[origin=r]{90}{MetalNut}}	& \multicolumn{1}{c}{\rotatebox[origin=r]{90}{Pill}}		
			  & \multicolumn{1}{c}{\rotatebox[origin=r]{90}{Screw}}			& \multicolumn{1}{c}{\rotatebox[origin=r]{90}{Tile}}		& \multicolumn{1}{c}{\rotatebox[origin=r]{90}{Toothbrush}} 		
			  & \multicolumn{1}{c}{\rotatebox[origin=r]{90}{Transistor}}	& \multicolumn{1}{c}{\rotatebox[origin=r]{90}{Wood}}		& \multicolumn{1}{c}{\rotatebox[origin=r]{90}{Zipper}}	
			  & \multicolumn{1}{c}{\rotatebox[origin=r]{90}{Average}}		\\ \hline\hline
			  
\multirow{6}{*}{\rotatebox[origin=c]{90}{$K=2$}} 	& TDG 					& 69.3	& 68.3	& 55.1	& 66.2	& 83.8	& 67.2	& 93.6	& 67.1	& 69.2	& 98.8	& 86.3	& 54.4	& 55.9	& 98.4	& 64.4	& 73.2\\ 
													& DifferNet				& 99.3	& 85.3	& 73.0	& 78.4	& 62.1	& 94.9	& 90.7	& 61.9	& 83.2	& 73.4	& 97.0	& 60.8	& 61.8	& 98.1	& 89.2	& 81.0\\ \cline{2-18}
													& RegAD 				& 99.4	& 65.1	& 67.5	& 96.5	& 84.0	& 96.0	& 99.4	& 91.4	& 81.3	& 52.5	& 94.3	& 86.6	& 86.0	& 99.2	& 86.3	& 85.7\\ 
													& + Ours			    & 99.8	& 65.9	& 70.2	& 96.9	& 77.0	& 96.3	& 100.0	& 94.9	& 80.7	& 66.0	& 99.4	& 83.2	& 82.6	& 99.7	& 86.6	& 86.6\\ \cline{2-18}
													& UniAD					& 99.9	& 60.1	& 65.7	& 100.0	& 90.6	& 90.4	& 100.0	& 63.0	& 62.5	& 75.3	& 99.4	& 91.4	& 67.9	& 98.1	& 90.6	& 83.7\\ 
													& + Ours			    & 100.0	& 58.5	& 64.7	& 99.9	& 94.3	& 91.9	& 100.0	& 64.6	& 63.0	& 73.1	& 99.2	& 91.1	& 68.4	& 97.9	& 91.1	& 83.8\\ \hline\hline
\multirow{6}{*}{\rotatebox[origin=c]{90}{$K=4$}}	& TDG 					& 69.6	& 70.3	& 47.6	& 68.7	& 86.2	& 71.2	& 93.2	& 69.2	& 64.7	& 98.8	& 87.2	& 57.8	& 67.7	& 98.3	& 65.3	& 74.4\\ 
													& DifferNet				& 99.3	& 85.2	& 80.3	& 78.6	& 60.5	& 95.8	& 91.2	& 67.3	& 84.0	& 72.5	& 98.0	& 62.5	& 62.2	& 96.4	& 84.8	& 81.0\\ \cline{2-18}
                                                    & RegAD 				& 99.4	& 76.1	& 72.4	& 97.9	& 91.2	& 95.8	& 100.0	& 94.6	& 80.8	& 56.6	& 95.5	& 90.9	& 85.2	& 98.6	& 88.5	& 88.2\\ 
                                                    & + Ours			& 99.6	& 77.0	& 77.5	& 98.5	& 83.4	& 96.6	& 100.0	& 94.3	& 85.9	& 60.2	& 99.2	& 91.2	& 85.0	& 99.6	& 91.5	& 89.3\\ \cline{2-18}
                                                    & UniAD					& 99.9	& 60.2	& 70.1	& 99.8	& 93.1	& 94.5	& 100.0	& 60.7	& 66.5	& 76.3	& 99.5	& 98.6	& 72.1	& 98.2	& 90.8	& 85.4\\ 
                                                    & + Ours			& 100.0	& 71.2	& 71.4	& 99.9	& 94.7	& 94.1	& 100.0	& 76.5	& 78.5	& 74.9	& 99.5	& 98.1	& 79.4	& 97.9	& 91.9	& 88.5\\ \hline\hline
\multirow{6}{*}{\rotatebox[origin=c]{90}{$K=8$}}	& TDG 					& 70.3	& 74.7	& 44.7	& 78.2	& 87.6	& 82.8	& 93.5	& 68.7	& 67.9	& 99.0	& 87.4	& 57.6	& 71.5	& 98.4	& 66.3	& 76.6\\ 
                                                    & DifferNet				& 99.4	& 87.9	& 78.6	& 78.5	& 78.5	& 97.9	& 92.2	& 67.7	& 82.1	& 75.0	& 99.6	& 60.8	& 63.3	& 99.4	& 87.3	& 83.0\\ \cline{2-18}
                                                    & RegAD 				& 99.8	& 80.6	& 76.3	& 98.5	& 91.5	& 96.5	& 100.0	& 98.3	& 80.6	& 63.4	& 97.4	& 98.5	& 93.4	& 99.4	& 94.0	& 91.2\\ 
                                                    & + Ours			& 99.9	& 85.1	& 80.6	& 96.7	& 87.3	& 96.8	& 100.0	& 94.5	& 84.4	& 70.1	& 99.9	& 98.7	& 90.9	& 99.2	& 94.7	& 91.9\\ \cline{2-18}
                                                    & UniAD					& 99.9	& 65.7	& 70.4	& 100.0	& 94.8	& 94.4	& 100.0	& 76.5	& 73.1	& 76.3	& 99.6	& 96.9	& 71.0	& 98.2	& 91.5	& 87.2\\ 
                                                    & + Ours			& 99.9	& 59.5	& 71.5	& 97.1	& 93.2	& 95.1	& 99.0	& 76.3	& 85.5	& 92.1	& 99.5	& 97.5	& 87.6	& 93.6	& 94.5	& 89.4\\ \hline\hline
\end{tabular}}
\caption{Quantitative evaluation of image level AUC on MVTec dataset \cite{mvtec}. Following the evaluation protocol of RegAD \cite{regad}, the image-level AUC scores of 10 runs are averaged. }
\label{tab:image_level}
\end{table*}

\subsection{Dataset}
The MVTec Anomaly Detection (MVTec AD) dataset \cite{mvtec} serves as a benchmark for anomaly detection, featuring a comprehensive and robust collection of 3629 high-resolution training images and 1725 testing images. 
The dataset includes normal and defective images, designed to resemble practical industrial inspection scenarios.
The dataset is categorized into textures, which display globally repetitive patterns, and objects, which adhere to predefined arrangements. 
With over 70 distinct defect types, the MVTec AD dataset challenges anomaly detection algorithms by accurately replicating real-world situations.
Additionally, the dataset provides pixel-precise ground truth (GT) regions, allowing for effective evaluation of detection techniques. 
Utilizing the MVTec AD dataset enables the assessment of our proposed method's performance in detecting and localizing various defect types across different categories, demonstrating its potential for real-world industrial applications.
Besides from MVTec AD dataset, we also experiment on the five classes of the DAGM2007 dataset \cite{dagm2007} to show the effectiveness of the proposed method.

\subsection{Experiments Detail}
In the experiments with both RegAD and UniAD, we extended the original models by incorporating a discriminator.
The discriminator consists of a sequence of convolutional layers for feature extraction, instance normalization layers for consistent scaling, and LeakyReLU activation functions for non-linear processing. 
Based on the pipeline of RegAD and UniAD, the same discriminator is used for experiments of the proposed method. 
The image size is resized to $224 \times 224$ for both models.
The training settings for RegAD consist of 50 epochs, a learning rate of 0.0001, and Stochastic Gradient Descent (SGD) with momentum, which are the same as the RegAD setting. 
On the other hand, UniAD employs 1000 epochs, a learning rate of 0.0001, an AdamW optimizer with betas of 0.9 and 0.999, and a weight decay of 0.0001, which are the same as the UniAD setting. 
For both RegAD and UniAD, the learning rate for the discriminator is set to the same as that for each model.
Following the few-shot setting of RegAD, each category of the few-shot target was trained using a leave-one-out approach for both RegAD and UniAD. 
This entails that the few-shot target category was exclusively used for testing, while the remaining categories were utilized for training.

\begin{table*}[t!]
\centering
\resizebox{\linewidth}{!}
{
\begin{tabular}{crrrrrrrrrrrrrrrr|r}
\hline\hline
&& \multicolumn{15}{c}{Category} \\ \cline{3-18}
\multicolumn{1}{c}{Shot} &\multicolumn{1}{c}{Method} & \multicolumn{1}{c}{\rotatebox[origin=r]{90}{Bottle}}		& \multicolumn{1}{c}{\rotatebox[origin=r]{90}{Cable}}		& \multicolumn{1}{c}{\rotatebox[origin=r]{90}{Capsule}}		
			  & \multicolumn{1}{c}{\rotatebox[origin=r]{90}{Carpet}}		& \multicolumn{1}{c}{\rotatebox[origin=r]{90}{Grid}}		& \multicolumn{1}{c}{\rotatebox[origin=r]{90}{Hazelnut}} 		
			  & \multicolumn{1}{c}{\rotatebox[origin=r]{90}{Leather}}		& \multicolumn{1}{c}{\rotatebox[origin=r]{90}{MetalNut}}	& \multicolumn{1}{c}{\rotatebox[origin=r]{90}{Pill}}		
			  & \multicolumn{1}{c}{\rotatebox[origin=r]{90}{Screw}}			& \multicolumn{1}{c}{\rotatebox[origin=r]{90}{Tile}}		& \multicolumn{1}{c}{\rotatebox[origin=r]{90}{Toothbrush}} 		
			  & \multicolumn{1}{c}{\rotatebox[origin=r]{90}{Transistor}}	& \multicolumn{1}{c}{\rotatebox[origin=r]{90}{Wood}}		& \multicolumn{1}{c}{\rotatebox[origin=r]{90}{Zipper}}	
			  & \multicolumn{1}{c}{\rotatebox[origin=r]{90}{Average}}		\\ \hline\hline
			  
\multirow{4}{*}{\rotatebox[origin=c]{90}{$K=2$}} 	& RegAD 				& 98.0	& 91.7	& 97.3	& 98.9	& 77.4	& 98.1	& 98.0	& 96.9	& 93.6	& 94.4	& 94.3	& 98.2	& 93.4	& 93.5	& 95.1	& 94.6\\ 
													& + Ours			& 98.6 & 93.9 & 97.5 & 98.9 & 80.0 & 98.4 & 99.4 & 97.8 & 97.8 & 94.8 & 96.3 & 96.6 & 94.3 & 96.8 & 97.4 & 95.9\\ \cline{2-18}
													& UniAD					& 95.4	& 83.9	& 95.4	& 99.6	& 92.8	& 95.0	& 99.0	& 72.4	& 82.7	& 91.4	& 90.6	& 96.0	& 81.7	& 93.3	& 94.3	& 90.9\\ 
													& + Ours			& 95.9	& 85.0	& 95.5	& 98.6	& 93.3	& 94.1	& 99.1	& 73.2	& 84.5	& 91.7	& 90.2	& 96.8	& 79.6	& 93.0	& 94.0	& 91.0\\ \hline\hline
\multirow{4}{*}{\rotatebox[origin=c]{90}{$K=4$}}	& RegAD 				& 98.4	& 92.7	& 97.6	& 98.9	& 85.7	& 98.0	& 99.1	& 97.8	& 97.4	& 95.0	& 94.9	& 98.5	& 93.8	& 94.7	& 94.0	& 95.8\\ 
                                                    & + Ours			& 98.6	& 96.1	& 98.3	& 98.9	& 83.0	& 98.7	& 99.5	& 96.8	& 97.8	& 96.3	& 95.7	& 97.9	& 93.8	& 96.6	& 97.6	& 96.4\\ \cline{2-18}
                                                    & UniAD					& 97.4	& 91.3	& 71.4	& 98.7	& 93.7	& 95.3	& 99.1	& 81.5	& 88.7	& 92.0	& 91.1	& 97.9	& 91.5	& 93.7	& 93.7	& 91.8\\ 
                                                    & + Ours			& 97.4	& 90.7	& 71.8	& 98.6	& 93.7	& 95.3	& 99.1	& 82.8	& 90.8	& 91.8	& 90.9	& 98.1	& 92.7	& 96.9	& 94.6	& 92.3\\ \hline\hline
\multirow{4}{*}{\rotatebox[origin=c]{90}{$K=8$}}	& RegAD 				& 97.5	& 94.9	& 98.2	& 98.9	& 88.7	& 98.5	& 98.9	& 96.9	& 97.8	& 97.1	& 95.2	& 98.7	& 96.8	& 94.6	& 97.4	& 96.7\\ 
                                                    & + Ours			& 98.5	& 96.8	& 98.4	& 98.8	& 86.2	& 98.8	& 99.2	& 98.0	& 98.1	& 97.4	& 96.2	& 98.9	& 96.5	& 94.9	& 96.7	& 96.9\\ \cline{2-18}
                                                    & UniAD					& 96.6	& 88.4	& 96.8	& 98.5	& 93.2	& 95.1	& 99.0	& 76.3	& 85.5	& 92.1	& 90.9	& 97.5	& 87.6	& 93.6	& 94.5	& 92.4\\ 
                                                    & + Ours			& 96.5	& 88.1	& 96.8	& 96.9	& 92.7	& 95.3	& 99.1	& 77.8	& 87.1	& 96.4	& 96.8	& 97.4	& 88.3	& 96.9	& 93.8	& 93.3\\ \hline\hline
\end{tabular}}
\caption{Quantitative evaluation of pixel level AUC on MVTec dataset \cite{mvtec}. Following the evaluation protocol of RegAD \cite{regad}, the pixel-level AUC scores of 10 runs are averaged.}
\label{tab:pixel_level}
\end{table*}

\begin{table*}[!t]
\centering
\resizebox{\linewidth}{!}
{
\begin{tabular}{crrrrrrr||rrrrrr}
\hline\hline
&& \multicolumn{6}{c}{Image-level} & \multicolumn{6}{c}{Pixel-level}\\ \cline{3-14}
\multicolumn{1}{c}{Shot} &\multicolumn{1}{c}{Method} & 
\multicolumn{1}{c}{Class1}		& \multicolumn{1}{c}{Class2}		& \multicolumn{1}{c}{Class3}		& \multicolumn{1}{c}{Class4}		& \multicolumn{1}{c}{Class5}		& \multicolumn{1}{c}{Average}		&
\multicolumn{1}{c}{Class1}		& \multicolumn{1}{c}{Class2}		& \multicolumn{1}{c}{Class3}		& \multicolumn{1}{c}{Class4}		& \multicolumn{1}{c}{Class5}		& \multicolumn{1}{c}{Average}		\\ \hline\hline
			  
\multirow{4}{*}{\rotatebox[origin=c]{90}{$K=2$}} 	& RegAD 			& 56.1	& 67.6	& 76.8	& 93.5	& 73.7	& 73.5	& 73.0   & 89.7	& 90.0 	& 97.7	& 82.1	& 86.5\\ 
													& + Ours			& 64.8	& 60.6	& 83.4	& 96.8	& 76.6	& 75.2	& 78.6	& 79.5	& 89.5	& 97.3	& 78.3	& 84.7\\ \cline{2-14}
													& UniAD				& 58.3	& 98.1	& 74.2	& 64.5	& 69.7	& 72.9	& 84.1	& 99.7	& 89.1	& 91.2	& 82.3	& 89.3\\ 
													& + Ours			& 60.0	& 98.0	& 74.8	& 66.4	& 70.6	& 74.0	& 84.1	& 99.8	& 88.9	& 91.9	& 82.9	& 89.5\\ \hline\hline
\multirow{4}{*}{\rotatebox[origin=c]{90}{$K=4$}}	& RegAD 			& 89.8	& 73.1	& 76.0	& 80.5	& 64.8	& 76.9	& 88.5	& 90.2	& 88.9	& 95.5	& 76.6	& 87.9\\ 
                                                    & + Ours			& 90.0	& 78.6	& 81.3	& 97.8	& 78.1	& 85.1	& 88.0	& 95.1	& 89.4	& 96.4	& 82.8	& 90.3\\ \cline{2-14}
                                                    & UniAD				& 59.4	& 98.1	& 74.1	& 78.6	& 70.4	& 76.1	& 85.0	& 99.8	& 88.8	& 93.7	& 82.3	& 89.9\\ 
                                                    & + Ours			& 59.1	& 98.1	& 75.8	& 79.4	& 70.9	& 76.7	& 85.1	& 99.8	& 89.6	& 94.0	& 82.5	& 90.2\\ \hline\hline
\multirow{4}{*}{\rotatebox[origin=c]{90}{$K=8$}}	& RegAD 			& 71.5  & 77.8  & 84.6  & 90.0  & 69.0  & 78.6  & 71.0  & 93.4  & 91.0  & 97.8  & 80.4  & 86.7\\ 
                                                    & + Ours			& 73.1  & 96.9  & 84.8  & 97.7  & 73.9  & 85.3  & 87.1  & 99.2  & 89.1  & 98.2  & 82.5  & 91.2\\ \cline{2-14}
                                                    & UniAD				& 59.1	& 98.1	& 75.7	& 88.3	& 71.0	& 78.4	& 85.1	& 99.8	& 89.9	& 95.5	& 82.7	& 90.6\\ 
                                                    & + Ours			& 59.3	& 98.0	& 76.1	& 93.2	& 72.2	& 79.8	& 85.3	& 99.8	& 90.0	& 96.2	& 83.1	& 90.9\\ \hline\hline
\end{tabular}}
\caption{Quantitative evaluation of image- and pixel-level AUC on DAGM2007 dataset \cite{dagm2007}. Following the evaluation protocol of RegAD \cite{regad}, the image- and pixel-level AUC scores of 10 runs are averaged. }
\label{tab:dagm}
\end{table*}

\begin{figure*}[!t]
    \centering
    \includegraphics[width=10cm]{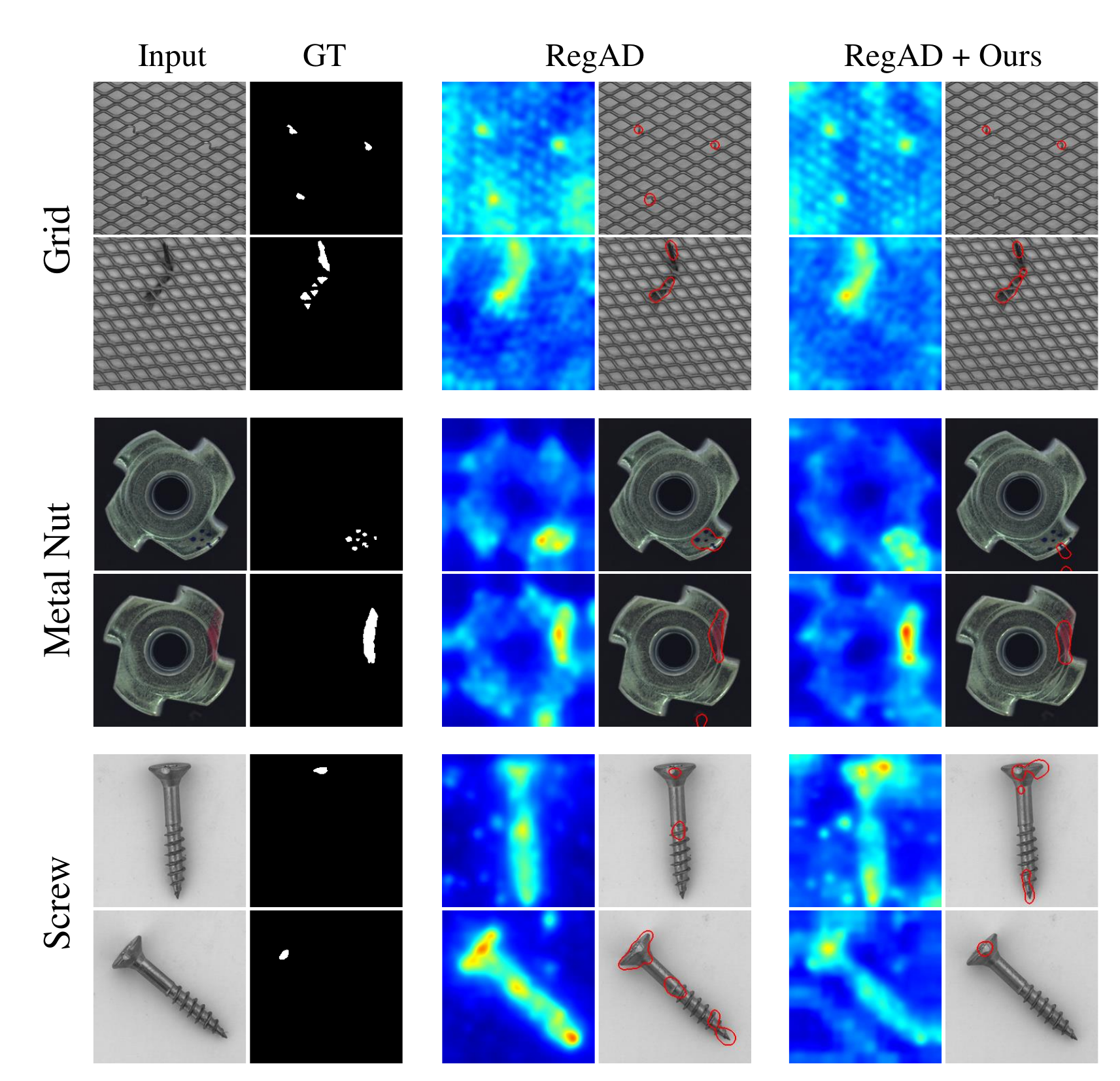}
    \caption{Qualitative comparison of 2-shot experiments between RegAD and RegAD + Ours on three categories: grid ($1^{st}$ and $2^{nd}$ rows), metal\_nut ($3^{rd}$ and $4^{th}$ rows), and screw ($5^{th}$ and $6^{th}$ rows). The input images and GTs are shown in the $1^{st}$ and $2^{nd}$ columns. The heat maps and prediction results of RegAD are shown in the $3^{rd}$ and $4^{th}$ columns. The heat maps and prediction results of RegAD + Ours are shown in the $5^{th}$ and $6^{th}$ columns.}
    \label{fig:qualitative_eval}
\end{figure*}

\subsection{Results and Discussions}
Table \ref{tab:image_level} shows a quantitative evaluation of image-level anomaly detection on the MVTec dataset. 
The results show the performance of various methods, including TDG \cite{TDG}, DifferNet \cite{differnet}, RegAD \cite{regad}, UniAD \cite{uniad}, and their variants combined with the proposed method (+ Ours). 
The performance is measured for different numbers of shots ($K=2$, $K=4$, and $K=8$) and across 15 different categories of objects in the MVTec AD dataset. 
The last column in each table shows the average performance across all categories.
In general, the proposed method, when combined with existing methods, tends to improve performance. 

As shown in Table \ref{tab:image_level}, when comparing RegAD and RegAD + Ours, the proposed method demonstrates better anomaly detection performance with an improvement ranging from 0.4 to 1.4 percentage points.
This improvement can be attributed to the proposed method's ability to better capture the distinctive features that differentiate normal and anomalous samples.
Similar to RegAD, when combined with the proposed method, UniAD also shows improvements ranging from 0.1 to 3.1 percentage points. 
This demonstrates the effectiveness of the proposed method for enhancing UniAD's performance. 

Table \ref{tab:pixel_level} presents a quantitative evaluation of pixel-level anomaly detection on the MVTec dataset, comparing the proposed method to RegAD and UniAD. 
Similar to the image-level evaluation, the proposed method shows general performance improvement when combined with existing methods for all $K$. 
This indicates the proposed method's ability to effectively enhance both image-level and pixel-level anomaly detection in various settings.

Table \ref{tab:dagm} shows the quantitative evaluation of RegAD, RegAD + Ours, UniAD, and UniAD + Ours on the DAGM2007 dataset.
Similar to the results of MVTec AD, while there may be a few classes where the proposed method performs lower than RegAD and UniAD, the overall experimental results with the proposed method show performance improvements in quantitative evaluation.

The results of the quantitative evaluation show that the proposed method helps to improve the performance of RegAD and UniAD, in general. 
However, the degree of improvement varies across different categories, suggesting a need for further investigation into the factors influencing the method's effectiveness.
Exploring the incorporation of complementary features or developing more sophisticated models could lead to better overall performance in anomaly detection, which we leave for future works.

\begin{figure*}[!t]
    \centering
    \includegraphics[width=9cm]{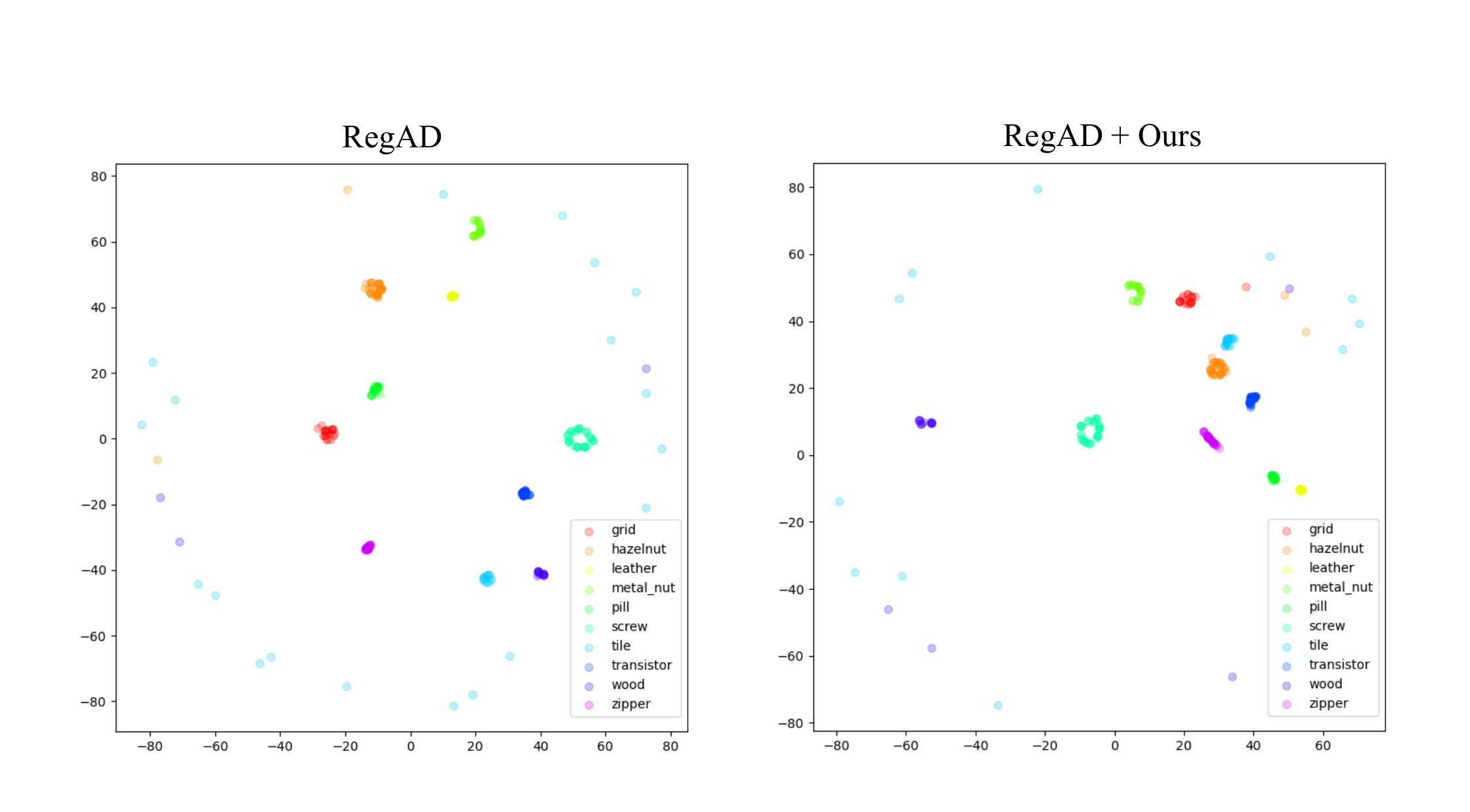}
    \caption{2-shot experiments: $t$-SNE visualization \cite{tsne} of features from normal samples using RegAD (left) and RegAD + Ours (right).}
    \label{fig:tsne}
\end{figure*}
\begin{figure*}[!t]
    \centering
    \includegraphics[width=12cm]{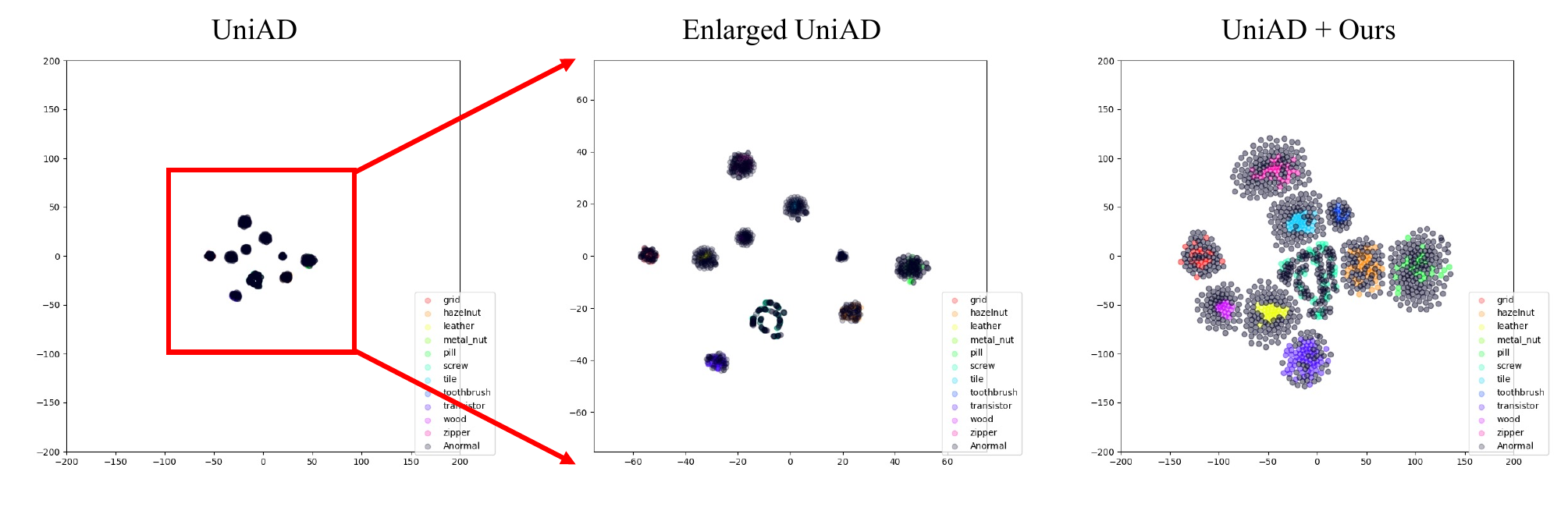}
    \caption{2-shot experiments: $t$-SNE visualization \cite{tsne} of features from normal and abnormal samples using UniAD (left), enlarged UniAD (middle), and UniAD + Ours (right).}
    \label{fig:tsne_uniad}
\end{figure*}

Figure \ref{fig:qualitative_eval} shows a qualitative comparison between RegAD and RegAD + Ours on three categories: grid, metal\_nut, and screw. 
As observed in the quantitative evaluation, the overall performance of the proposed method improves. 
However, there are categories, such as grid, where performance is lower than that of RegAD across 2, 4, and 8 shots. 
Although a large number of samples appear visually better for RegAD, as shown in the first row of Figure \ref{fig:qualitative_eval}, there are some samples where the proposed method shows visually better performance (the $2^{nd}$ row). 
Conversely, in categories like metal\_nut and screw, even though the quantitative evaluation indicates better performance for the proposed method, some samples exhibit visually better results for RegAD, as shown in the $3^{rd}$ and $5^{th}$ rows of Figure \ref{fig:qualitative_eval}.
In supplementary material, qualitative evaluation of more samples is presented.

Figure \ref{fig:tsne} shows the $t$-SNE visualization of features extracted from normal samples using RegAD \& RegAD + Ours and UniAD \& UniAD + Ours, respectively. 
Ideally, normal samples should form distinct category-wise clusters.
However, the visualizations for both RegAD and RegAD + Ours reveal that some samples have considerable distances from their respective category clusters. 
These samples are incorrectly identified as abnormal samples. 
Notably, the visualization demonstrates that the number of incorrectly predicted samples in RegAD + Ours is smaller than that of RegAD (18 samples for RegAD + Ours and 24 samples for RegAD). 
It indicates that the proposed method effectively contributes to clustering the features of normal samples. 

Similarly, Figure \ref{fig:tsne_uniad} shows $t$-SNE \cite{tsne} visualization of features extracted from normal and abnormal samples using UniAD \cite{uniad} and UniAD + Ours on 2-shot experimental setting. 
Each sub-figure from leftmost to rightmost figures in Figure \ref{fig:tsne_uniad} represents $t$-SNE visualization of UniAD, enlarged version of UniAD, and UniAD + Ours, respectively. 
In UniAD, although the features are clustered, the feature space appears to be narrow. 
However, in UniAD + Ours, the scale of the feature space is enlarged while preserving the feature clusters. 
Despite the enlargement of the feature space in UniAD, the normal and abnormal samples are not well-distinguished. 
Compared to the visualization of UniAD, UniAD + Ours shows that the abnormal samples are relatively well-distinguished from normal samples. 
These $t$-SNE visualizations show that the proposed method is effective in improving the distinction between normal and abnormal samples.

\section{Conclusion}
In this paper, we propose a FSAD method by incorporating the adversarial loss, commonly used in domain adaptation.
The proposed method leverages the power of Siamese networks and adversarial training to enhance the generalization capability of existing anomaly detection models. 
By incorporating adversarial loss into the training process, we encouraged the model to generate more robust and generalized features that aid in better distinguishing normal and anomalous samples.
We experimented using MVTec AD and DAGM datasets and demonstrated the effectiveness of the proposed method in combination with existing FSAD methods, such as RegAD and UniAD. 
The proposed method consistently improved the performance of these models in both image-level and pixel-level evaluations.
However, it is noteworthy that further investigation to assess factors influencing detection performance is necessary because the level of improvement varies across categories. 
For future work, we will conduct a further study to explore how to achieve a stable performance enhancement by developing more sophisticated models.

\section{Acknowledgement}
This work was supported by Institute of Information \& communications Technology Planning \& Evaluation (IITP) grant funded by the Korea government(MSIT) (No.2022-0-00866, Development of cyber-physical manufacturing base technology that supports high-fidelity and distributed simulation for large-scalability)

\vspace{1em}
\noindent The authors deeply appreciate Kang Il Choi for kindly sharing computing resources with us.
\bibliography{egbib}
\end{document}